\documentclass{article}

    \PassOptionsToPackage{numbers, compress}{natbib}

 \usepackage[preprint]{neurips_2026}

\usepackage[utf8]{inputenc} %
\usepackage[T1]{fontenc}    %
\usepackage{url}            %
\usepackage{booktabs}       %
\usepackage{amsfonts}       %
\usepackage{nicefrac}       %
\usepackage[protrusion=true, expansion=false]{microtype}

\usepackage{graphicx}      %
\usepackage{multirow}      %
\usepackage[table]{xcolor} %
\usepackage{amsmath}
\usepackage{enumitem}
\usepackage{wrapfig}

\definecolor{iccvblue}{rgb}{0.21,0.49,0.74}
\usepackage[pagebackref,breaklinks,colorlinks,allcolors=iccvblue]{hyperref}
\usepackage[nameinlink,capitalize]{cleveref}

\crefname{section}{Sec.}{Secs.}
\Crefname{section}{Section}{Sections}
\Crefname{table}{Table}{Tables}
\crefname{table}{Tab.}{Tabs.}

\usepackage{capt-of}
\usepackage{caption}

\usepackage[normalem]{ulem}

\hypersetup{
  colorlinks=true,
  urlcolor=iccvblue
}
\urldef{\projecturl}\url{https://github.com/Tianhang-Wang/DecQ}

\title{DecQ: Detail-Condensing Queries for Enhanced Reconstruction and Generation \\ in Representation Autoencoders}

\author{%
  \textbf{Tianhang Wang$^{1,2,*}$} \quad
  \textbf{Yitong Chen$^{2,3,*}$} \quad
  \textbf{Wei Song$^{1,2,4}$} \\[-0em]
  \textbf{Zuxuan Wu$^{2,3,\dagger}$} \quad
  \textbf{Min Li$^{1,\dagger}$} \quad
  \textbf{Jiaqi Wang$^{2,5,\dagger}$} \\
  \\[-0.15em]
  $^{1}$Zhejiang University \quad
  $^{2}$Shanghai Innovation Institute \quad
  $^{3}$Fudan University \\[-0em]
  $^{4}$Westlake University \quad
  $^{5}$JD.COM\\
  \\[-0.15em]
  \projecturl
}

\begin{document}

\maketitle
\begingroup
\renewcommand{\thefootnote}{\fnsymbol{footnote}}
\footnotetext[1]{Equal contribution.}
\footnotetext[2]{Corresponding authors.}
\endgroup

\vspace{-15pt}
\begin{abstract}
\vspace{-5pt}

Representation Autoencoders (RAEs) leverage frozen vision foundation models (VFMs) as tokenizer encoders, providing robust high-level representations that facilitate fast convergence and high-quality generation in latent diffusion models. However, freezing the VFM inherently constrains its spatial reconstruction capacity, limiting fine-grained generation and image editing; in contrast, incorporating reconstruction-oriented signals via fine-tuning disrupts the pretrained semantic space and degrades generative fidelity.
To address this trade-off, we propose \textbf{DecQ}, a simple yet effective framework for RAEs. Specifically, DecQ introduces lightweight detail-condensing queries that extract fine-grained information from intermediate VFM features through condenser modules. These queries are incorporated into the decoder to support reconstruction and are jointly generated with patch tokens during generative modeling. By aggregating information from both shallow and deep layers, DecQ effectively mitigates the reconstruction--generation trade-off, improving both reconstruction quality and generative performance.
Our experiments demonstrate that: (1) with only 8 additional queries and 3.9\% extra computation, DecQ improves reconstruction over the frozen DINOv2-based RAE, increasing PSNR from 19.13 dB to 22.76 dB; and (2) for generative modeling, DecQ achieves 3.3$\times$ faster convergence than RAE, attaining an FID of 1.41 without guidance and 1.05 with guidance.

\vspace{-5pt}
\end{abstract}

\begin{figure}[ht]
  \centering

  \begin{minipage}[t]{0.59\textwidth}
    \centering
    \includegraphics[width=\linewidth]{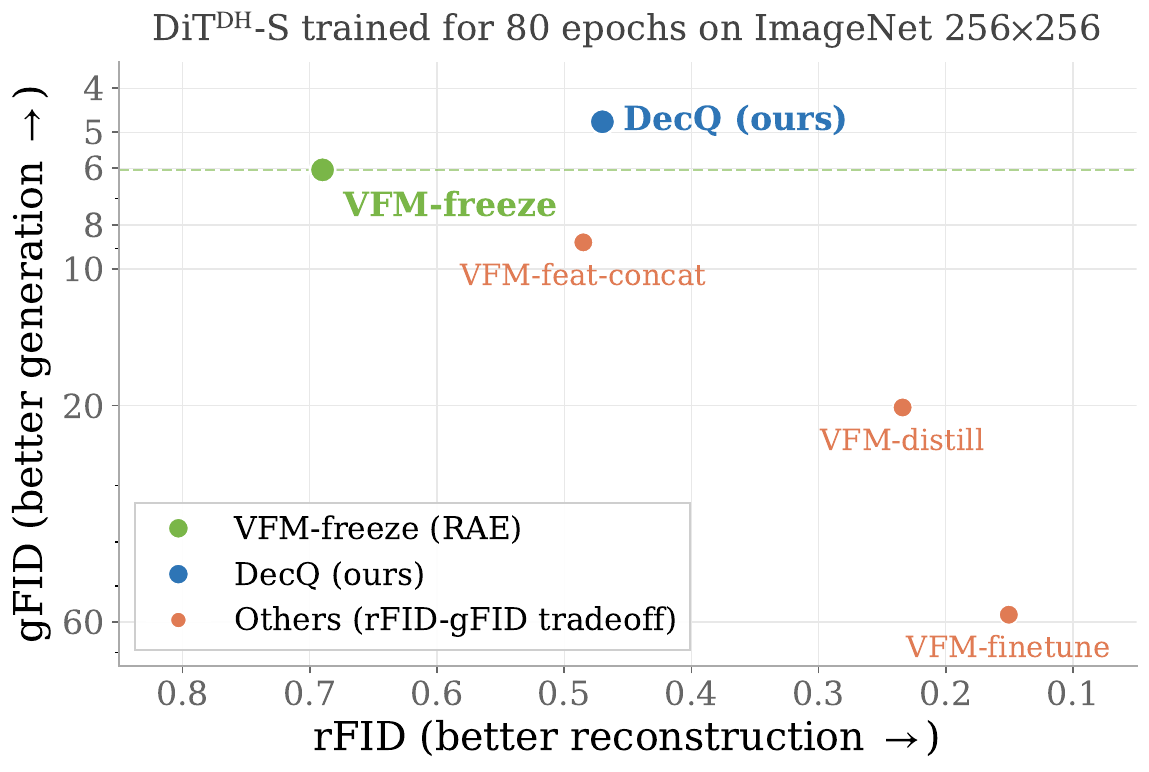}
  \end{minipage}
  \hfill
  \begin{minipage}[t]{0.4\textwidth}
    \centering
    \includegraphics[width=\linewidth]{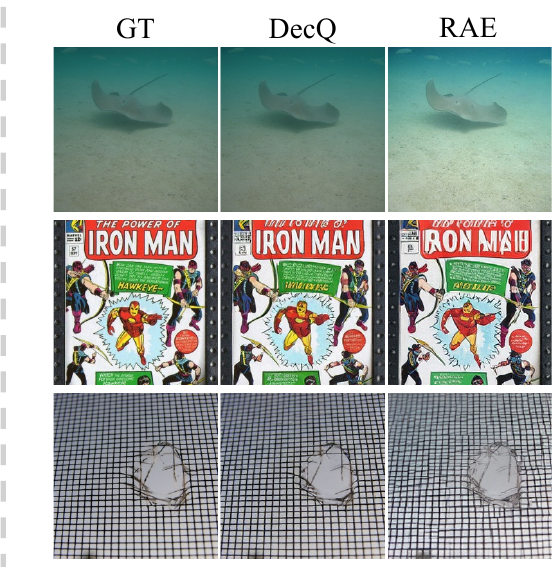}
  \end{minipage}

  \caption{
        \textbf{(Left) An empirical study of different VFM-based image tokenizer paradigms based on DINOv2.}
        \textit{VFM-freeze}, corresponding to the RAE baseline, keeps the VFM encoder frozen and directly uses its representations for reconstruction.
        \textit{VFM-finetune} denotes directly fine-tuning the VFM encoder, \textit{VFM-distill} uses a frozen VFM copy to distill the encoder outputs, and \textit{VFM-feat-concat} keeps the VFM frozen while concatenating low-level information along the feature dimension.
        These variants improve reconstruction quality but reveal a clear trade-off between rFID and gFID, leading to degraded generative performance.
        In contrast, DecQ improves reconstruction while also enhancing generation compared with RAE, demonstrating that detail-condensing queries can enrich low-level details without compromising the VFM semantic latent space.
        Detailed results are provided in \cref{tab:backbone_ablation}.
        \textbf{(Right) Reconstruction examples of our DecQ framework.}
        Our reconstruction preserves more color and texture information than RAE, enabling more accurate recovery of background colors and better reconstruction of textual content and fine-grained textures.
  }
  \vspace{-3mm}
  \label{fig:teaser}
\end{figure}

\vspace{-4pt}
\section{Introduction}
\label{sec:intro}

{In visual generation, state-of-the-art diffusion models~\cite{LDM,flux,sana} are typically built upon a two-stage training paradigm: first learning a tokenizer, and then training a generative model in the resulting latent space, where the tokenizer is usually implemented as an autoencoder. Recently, Representation autoencoders (RAEs)~\cite{rae} revisit this design by replacing the tokenizer encoder with frozen pretrained vision foundation models (VFMs)~\cite{dinov2,siglip2} and training only an additional decoder. Their results demonstrate that the semantically rich latent space induced by such pretrained representations can substantially accelerate the convergence of diffusion models.}

Despite their advantages, directly using frozen VFMs as image tokenizers introduces a clear objective mismatch. Existing VFMs are typically trained with multimodal alignment~\cite{CLIP,siglip,siglip2} or self-distillation~\cite{dinov2,dinov3} objectives, rather than explicit pixel-level reconstruction losses~\cite{vae,vq-vae,vqgan}. These objectives often encourage invariance across augmented views~\cite{dinov2,siglip2}, which improves semantic robustness but may reduce sensitivity to low-level cues such as color and texture~\cite{unilip,svg,dualtoken}. As a result, frozen VFM latent representations are not well suited to serve as information-preserving image codes. When used as frozen encoders, their limited preservation of low-level details can lead to reconstruction artifacts such as texture loss and color shifts, as shown in~\cref{fig:teaser} Right. Due to the limited invertibility of such frozen representations, RAE models built upon frozen VFMs may exhibit weaker fine-grained generation and editing capabilities. In other words, although RAEs often enable faster convergence, their ultimate generative performance can be substantially constrained by suboptimal reconstruction fidelity.

To address this challenge, a straightforward strategy is to inject more low-level, reconstruction-oriented features into the latent space. Prior works~\cite{unilip,aligntok,rpiae} explore fine-tuning VFMs on reconstruction tasks while introducing a semantic distillation loss to preserve the original VFM outputs. However, such designs impose conflicting objectives, leading to an inherent trade-off between semantic consistency and reconstruction fidelity.
Other approaches~\cite{lvrae,svg} instead directly augment the latent space with reconstruction-relevant information. Nevertheless, these methods also inject low-level signals that can interfere with the original semantic representations, potentially hindering the convergence of downstream generative models.

For a more controlled and fair comparison, we conduct an empirical study of different VFM-based image tokenizer paradigms under a unified setting in \cref{fig:teaser} (Left), where all methods use DiT$^{\mathrm{DH}}$-S as the generative model and are trained on ImageNet at $256 \times 256$ resolution for 80 epochs. Specifically, \emph{VFM-finetune} directly unfreezes the VFM encoder during training; \emph{VFM-distill} trains the encoder with an additional distillation loss from a frozen VFM teacher; and \emph{VFM-feat-concat} freezes the VFMs while augmenting reconstruction information through feature-dimensional concatenation. Despite their differences, all variants exhibit a consistent reconstruction--generation trade-off: improved reconstruction fidelity comes at the cost of degraded generative performance.

In this paper, we propose DecQ, a framework designed to resolve this dilemma. DecQ introduces a small set of learnable queries that attend to the intermediate features of a frozen VFM, forming detail-condensing queries that capture low-level reconstruction details complementary to the semantic latent space. Since the VFM remains frozen, these queries enrich fine-grained details without modifying the original VFM parameters or perturbing its semantic representations. DecQ further incorporates these queries into the generative process by jointly denoising them with image patches. We find that predicting the detail-condensing queries also benefits generation, mitigating the reconstruction–generation trade-off. Our main contributions are summarized as follows:

\begin{itemize}[leftmargin=1.2em, labelsep=0.5em, itemsep=0.25em, topsep=0.3em, parsep=0pt]

    \item We propose DecQ, a representation autoencoder framework that uses a small set of learnable queries to capture low-level details under-represented by VFMs via cross-attention. It improves fine-grained reconstruction without changing the original pretrained VFM latent space.
    \item We find that condensing features from shallow VFM layers mainly benefits reconstruction, while condensing features from deep VFM layers benefits generation. By condensing information from both shallow and deep VFM layers, these queries effectively improve reconstruction and generation simultaneously.
    \item Extensive experiments demonstrate that DecQ improves reconstruction over the frozen-VFM baseline while consistently benefiting generative performance, achieving faster convergence and better generation quality with only limited additional overhead.

\end{itemize}

\vspace{-4pt}
\section{Related work}
\vspace{-4pt}
\label{gen_inst}

\textbf{Representation Alignment in Diffusion Models.~}
Latent Diffusion Models (LDMs) based on Diffusion Transformers (DiTs) have received increasing attention~\cite{LDM,dit,SiT}. However, vanilla DiTs often suffer from slow convergence and limited generation performance. To accelerate DiT training, REPA~\cite{repa} aligns the noisy hidden states of the diffusion model with clean representations from VFMs. Subsequent works~\cite{irepa,repa-e,reg,catok2026} further improve this framework from several complementary directions. iREPA~\cite{irepa} refines the alignment mechanism, showing that spatial structure is more crucial than global semantics and enhancing feature transfer via spatial normalization. REPA-E~\cite{repa-e} leverages the representation alignment objective to unlock the end-to-end joint tuning of the VAE and DiT without causing latent space collapse. Furthermore, REG~\cite{reg} addresses the absence of alignment during inference by jointly denoising image latents and a VFM class token, providing continuous semantic guidance for better generation fidelity.

\textbf{VFM-Aligned Visual Tokenizers for Generation.~}
From another perspective, several works focus on improving the visual tokenizer itself, arguing that its latent space should inherently possess strong semantics~\cite{va-vae,aligntok,dmvae,veloss}. For instance, VA-VAE~\cite{va-vae} directly aligns the VAE latent space with pretrained foundation models. Similarly, AlignTok~\cite{aligntok} aligns a pretrained VFM to a visual tokenizer rather than forcing the encoder to learn semantics from scratch. Furthermore, DMVAE~\cite{dmvae} leverages Distribution Matching Distillation (DMD) to explicitly constrain the encoder’s aggregate posterior to match a predefined reference distribution, such as a self-supervised learning prior. %

\textbf{VFMs as Direct Tokenizers for Generation.~}
Recent works, particularly RAE, introduce the idea of directly adopting VFMs as latent encoders for LDMs, enabling generation in the high-dimensional semantic latent space of VFMs with techniques such as noise shift and the DDT head~\cite{rae,vfmvae,ddt,sd3}. Benefiting from the strong semantics of VFMs, RAE achieves faster convergence and improved generation performance. Concurrently, SVG~\cite{svg} improves reconstruction in VFM-based latent spaces by concatenating additional reconstruction-oriented information along the feature dimension.
Subsequent works~\cite{fae,rpiae,lvrae} have improved upon RAE. For instance, FAE~\cite{fae} uses a semantic autoencoder to compress the VFM latent space into a lower-dimensional latent space for more efficient generation. Unlike FAE, which completely freezes the encoder, RPiAE~\cite{rpiae} proposes a multi-stage training process that initializes from the VFM but allows fine-tuning for reconstruction. To maintain pixel-wise reconstruction quality, LVRAE~\cite{lvrae} adds the low-level information under-represented by VFMs back into the output space.
However, these methods generally modify or reshape the original VFM semantic space. LVRAE introduces additional low-level information into the output representation, while FAE and RPiAE compress the semantic space into lower dimensions; RPiAE further changes the representation by fine-tuning the VFM itself.
In contrast, DecQ preserves the original VFM semantic space. By introducing detail-condensing queries that capture reconstruction-oriented details from VFMs, DecQ simultaneously improves reconstruction fidelity and generation performance with limited extra overhead.

\vspace{-4pt}
\section{Method}
\vspace{-4pt}
\label{headings}

In this section, we first review the preliminaries of representation autoencoders in \cref{sec:preliminary}. We then introduce the tokenizer training procedure of our DecQ framework in \cref{sec:decq}. Finally, we outline the diffusion modeling used for image generation with the trained DecQ tokenizer in \cref{sec:decq_gen}.

\subsection{Preliminary}
\label{sec:preliminary}

The standard paradigm for Diffusion Transformers typically relies on a compressed latent space defined by a Variational Autoencoder (VAE). However, the reconstruction-centric objective of VAEs often results in representations that are less semantically structured. This motivates the RAE \cite{rae} framework, which redefines latent generative modeling by leveraging frozen, semantically-rich VFMs as the latent space. In RAE, a frozen VFM encoder $E$ extracts high-dimensional latent tokens $z = E(x)$, while a ViT-based decoder $D$ reconstructs images using a combination of pixel-wise ($L_1$), perceptual (LPIPS), and adversarial (GAN) losses \cite{lpips,StyleGAN-T,vit}. To model this high-dimensional space, RAE adopts a flow matching formulation that interpolates between the latent distribution $p(z)$ and Gaussian noise $\mathcal{N}(0, I)$: $z_t = (1-t)z + t\epsilon$ for $t \in [0, 1]$. A Diffusion Transformer $v_\theta$ is then trained to approximate the optimal velocity field $v(z_t, t) = \mathbb{E}[\epsilon - z | z_t]$ by minimizing the mean-squared error objective:
\begin{equation}
\mathcal{L}_{velocity}(\theta) = \int_{0}^{1} \mathbb{E}_{z, \epsilon} \left[ \Vert v_\theta(z_t, t, y) - (\epsilon - z) \Vert^2 \right] dt,
\end{equation}
where $y$ represents optional class-conditional information.

Despite RAE's effectiveness in capturing high-level semantics, its tokenizer has a key limitation: its latent space consists entirely of patch tokens from a VFM encoder, which are naturally biased toward semantic abstraction. While these tokens encode global semantics well, they under-represent low-level visual details essential for faithful reconstruction, such as color fidelity and fine-grained textures. This motivates a mechanism that supplements fine-grained low-level information while preserving the frozen VFM latent space.

\subsection{DecQ Tokenizer}
\label{sec:decq}
\begin{figure*}[t]
  \centering
  \includegraphics[width=\textwidth]{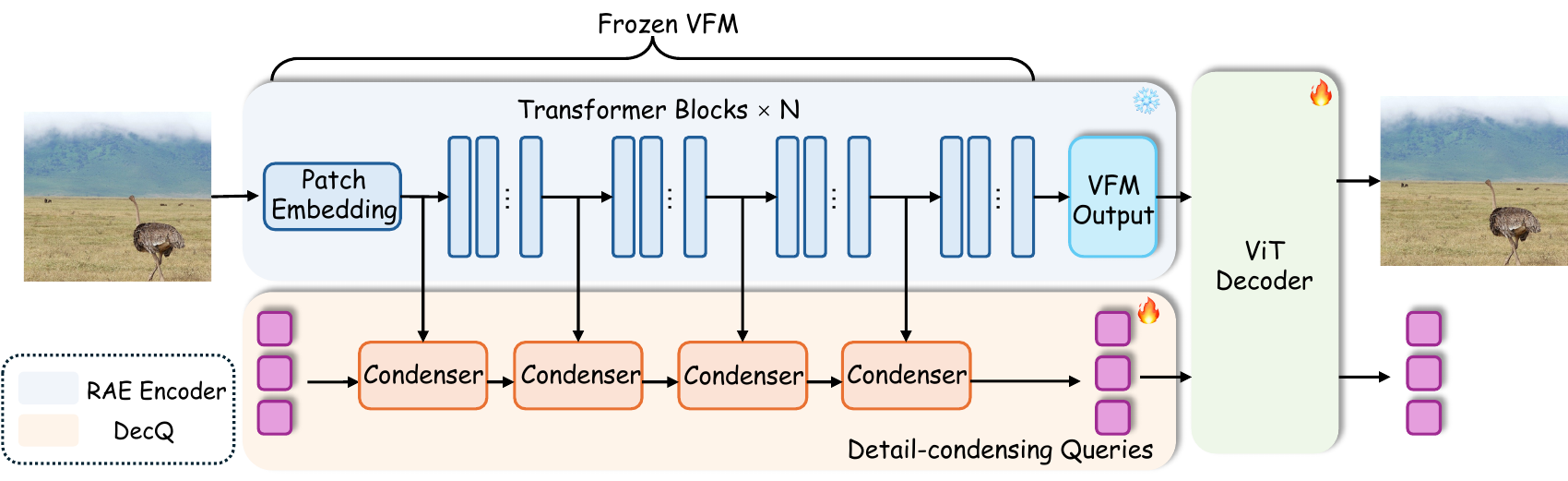}
  \caption{
        \textbf{Overview of the DecQ architecture.} Given an input image, the frozen VFM first converts it into patch tokens and processes them through a stack of Transformer blocks. DecQ attaches learnable queries to multiple intermediate VFM layers and uses condenser modules to progressively aggregate multi-level features into detail-condensing queries. These queries are then fed into the ViT decoder together with the VFM output tokens, providing complementary fine-grained details while keeping the VFM semantic space unchanged.
  }
  \label{fig:arch}
\end{figure*}

To address this limitation, we introduce \emph{DecQ}, a lightweight tokenizer extension that augments frozen VFM patch tokens with detail-condensing queries. These queries condense complementary low-level information from intermediate layers of the frozen encoder, improving reconstruction quality with minimal additional cost. An overview of DecQ is shown in \cref{fig:arch}.

\begin{wrapfigure}{r}{0.28\textwidth} %
    \centering
    \includegraphics[width=\linewidth]{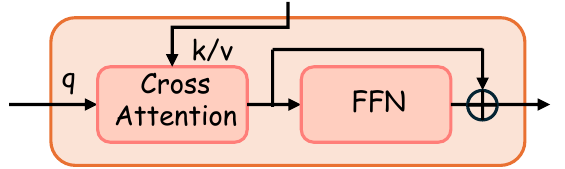} %
    \caption{\textbf{Architecture of the condenser.}} %
    \label{fig:encoder_condensers}
\end{wrapfigure}
\paragraph{Encoder with Condensers.}%
We introduce $K$ learnable query tokens $\mathbf{Q}^{(0)} \in \mathbb{R}^{K \times C}$ alongside the frozen VFM backbone, where $C$ is the feature dimension of the patch tokens. In practice, $K \ll N$, so the query tokens provide a compact representation for complementary fine-grained information. To aggregate multi-level features without modifying the pretrained VFM representations, we attach condenser modules to intermediate layers of the frozen encoder. As shown in \cref{fig:encoder_condensers}, each condenser consists of a cross-attention block followed by an FFN. In the cross-attention block, the query tokens serve as queries, while the intermediate patch tokens serve as keys and values. Let $\mathbf{Q} \in \mathbb{R}^{K \times C}$ and $\mathbf{P} \in \mathbb{R}^{N \times C}$ denote the query and patch tokens, respectively. The cross-attention is defined as:
\begin{equation}
\mathrm{CrossAttn}(\mathbf{Q}, \mathbf{P}) = \mathrm{Softmax}\left(\frac{\mathbf{Q}W_Q (\mathbf{P}W_K)^\top}{\sqrt{d}}\right)\mathbf{P}W_V,
\end{equation}
where $W_Q, W_K, W_V$ are learnable projection matrices, and $d$ denotes the attention head dimension.
At layer $l$, the query tokens condense information from the intermediate VFM patch tokens $\mathbf{P}^{(l)}$ through a residual cross-attention block followed by an FFN:
\begin{align}
    \tilde{\mathbf{Q}}^{(l)} &= \mathbf{Q}^{(l)} + \mathrm{CrossAttn}\big(\mathrm{LN}(\mathbf{Q}^{(l)}), \mathrm{LN}(\mathbf{P}^{(l)})\big), \\
    \mathbf{Q}^{(l+1)} &= \tilde{\mathbf{Q}}^{(l)} + \mathrm{FFN}\big(\mathrm{LN}(\tilde{\mathbf{Q}}^{(l)})\big).
\end{align}

Since patch tokens are only used as keys and values, information flows from patches to queries. This unidirectional design prevents query tokens from altering the pretrained VFM representations, thereby preserving the original semantic latent space. The encoder outputs two types of latents: semantic patch tokens $\mathbf{Z}_{\text{patch}}$ and detail-condensing query tokens $\mathbf{Z}_{\text{query}}$.

\paragraph{Dual-Stream Decoder.}
We follow the ViT decoder recipe of RAE and incorporate both patch and query tokens. Patch and query tokens are first projected to the decoder dimension using separate linear layers. We add fixed 2D sinusoidal positional embeddings to the patch tokens and learnable positional embeddings to the query tokens. The two token sequences are then concatenated:
\begin{equation}
\mathbf{H}^{(0)} = \left[ \mathbf{Z}_{\text{patch}} + \mathbf{PE}_{\text{2D}} \;\|\; \mathbf{Z}_{\text{query}} + \mathbf{PE}_{Q} \right],
\end{equation}
where $[\cdot \| \cdot]$ denotes concatenation. 
The combined sequence is processed jointly by the decoder. Only patch tokens are used for pixel prediction, while query tokens participate in decoder self-attention to provide fine-grained details. Finally, following the regularization strategy of RAE, we apply noise augmentation to both patch and query latents during training.

\subsection{Generation with Detail-Condensing Queries}
\label{sec:decq_gen}
\begin{figure*}[t]
  \centering
  \includegraphics[width=\textwidth]{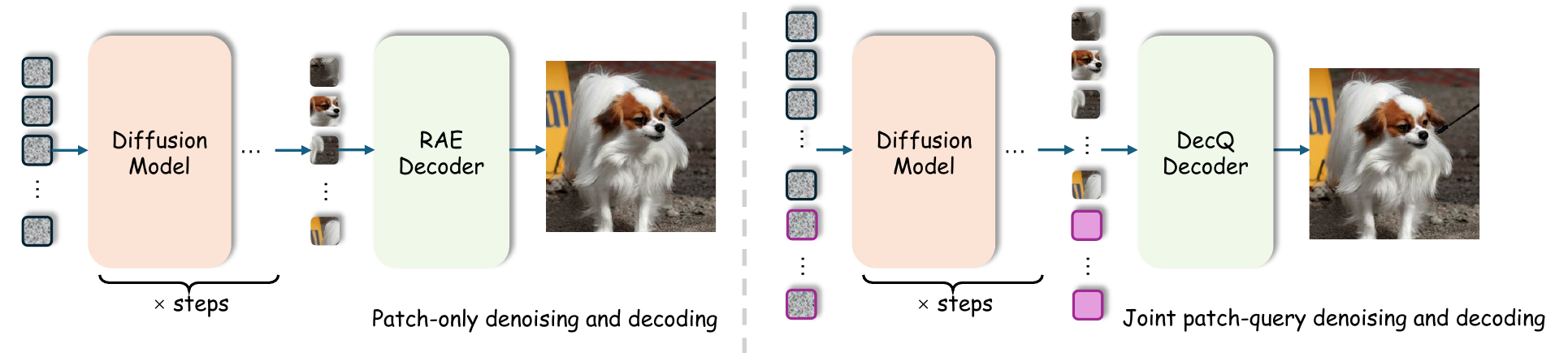}
  \caption{
        \textbf{Image generation with detail-condensing queries.} 
        Compared with the RAE baseline that denoises and decodes only VFM patch tokens, DecQ jointly denoises detail-condensing queries with patch tokens during diffusion. 
        Both patch and query tokens are initialized from Gaussian noise and generated as a unified latent sequence, and are then jointly decoded into the output image.
  }
  \label{fig:gen_arch}
\end{figure*}

As shown in \cref{fig:gen_arch}, in the generation stage, we extend the latent space by concatenating semantic patch tokens and detail-condensing query tokens into a single sequence:
$\mathbf{Z} = [\mathbf{Z}_{\text{patch}} \,\|\, \mathbf{Z}_{\text{query}}].$
This extended sequence preserves the semantic structure of the frozen VFM patch tokens while incorporating complementary fine-grained details from the query tokens. During generative modeling, patch and query tokens are jointly denoised and then fed into the decoder to decode the output image.

\paragraph{Sequence Modeling.}
We model the extended latent sequence $\mathbf{Z}$ using the DiT$^{\mathrm{DH}}$ architecture adopted in RAE \cite{rae,ddt}, trained under the flow matching objective. Patch and query tokens are jointly denoised with global self-attention and are then jointly fed into the decoder to produce the final image. To account for their different token types, we use separate input projections and positional encodings: patch tokens are equipped with 2D positional embeddings, while query tokens use independent learnable positional embeddings.

\paragraph{Optimization and Inference.} 
During training, the flow matching velocity prediction loss is computed over the full sequence and decomposed as
\begin{equation}
    \mathcal{L} = \mathcal{L}_{\text{patch}} + \lambda_{\text{query}} \cdot \mathcal{L}_{\text{query}},
\end{equation}
where $\mathcal{L}_{\text{patch}}$ and $\mathcal{L}_{\text{query}}$ denote the mean squared error (MSE) over patch and query tokens, respectively, and $\lambda_{\text{query}}$ controls the weight of query-token prediction. At inference time, we sample Gaussian noise for the full latent sequence and integrate the flow ODE to obtain both patch and query latents, which are then decoded into the final image.

\section{Experiments}
\label{sec:exp}

\subsection{Experimental Settings}
\label{sec:exp_settings}

We follow the RAE experimental protocol and keep key generation settings, including the dimension-dependent time shift and wide DDT head~\cite{sd3,ddt}, consistent with the original RAE configuration. Unless otherwise specified, we use DINOv2-B as the default VFM and a ViT-XL decoder with approximately 500M parameters, and conduct experiments on ImageNet~\cite{imagenet1k} at $256 \times 256$ resolution. By default, DecQ uses 8 detail-condensing queries, with condensers attached to VFM layers 0, 3, 6, and 9. During diffusion training, query and patch tokens share the same noise schedule, and the query-token loss weight is set to 1. Additional details are provided in Appendix~\ref{sec:implement}.

For reconstruction, we report PSNR and SSIM~\cite{ssim} for pixel-level fidelity, and Fr\'echet Inception Distance (FID)~\cite{fid}, denoted as rFID, for distributional quality and visual realism. %
For generation, we report FID, Inception Score (IS), Precision (Prec.), and Recall (Rec.), with generation FID denoted as gFID. Metrics are computed using the ADM evaluation suite on 50,000 class-uniform samples~\cite{adm}. Unless otherwise specified, we use 50 sampling steps following the RAE protocol.

\subsection{Main Results}
\label{sec:main_results}

\subsubsection{Reconstruction Ability}

\noindent
\begin{minipage}[c]{0.52\textwidth}
    We report reconstruction results in \cref{tab:reconstruction_main}. Among VFM-based tokenizers, DecQ achieves the best rFID, while also substantially improving pixel-level reconstruction metrics over the original RAE at a resolution of $256\times256$. These gains indicate that DecQ recovers significantly richer low-level visual details while faithfully preserving the high-level semantic structure of the latent space. 
    Notably, DecQ does not introduce any additional encoder to extract information directly from the input image. Instead, it leverages intermediate features within the frozen VFM to recover fine-grained information that is progressively lost along the forward pass. This design is both lightweight and structurally consistent with the original representation space. Additional qualitative results are provided in Appendix~\ref{sec:qualitative}.
\end{minipage}
\hfill
\begin{minipage}[c]{0.47\textwidth}
    \centering
    \captionsetup{type=table}
    \caption{\textbf{Quantitative comparison of reconstruction performance across different VFM-based tokenizers}. Our proposed DecQ achieves the lowest rFID within VFM-based encoders and significantly outperforms RAE in pixel-wise reconstruction metrics.}
    \label{tab:reconstruction_main}
    \small
    \begin{tabular}{l ccc}
    \toprule
    \textbf{Method} & \textbf{rFID$\downarrow$} & \textbf{SSIM$\uparrow$} & \textbf{PSNR$\uparrow$} \\
    \midrule
    SD-VAE \cite{LDM}                              & 0.61 & 0.74 & 26.90 \\
    VA-VAE \cite{va-vae}                           & 0.28 & 0.79 & 27.96 \\
    \midrule
    SVG \cite{svg}                               & 0.65 & \textbf{0.65}  & \textbf{23.89}  \\
    RAE \cite{rae}                     & 0.69 & 0.49 & 19.13 \\
    RPiAE \cite{rpiae}                             & 0.50 & 0.53 & 21.30 \\
    FAE \cite{fae}
    & 0.68 & --    & --    \\
    DecQ               & \textbf{0.47} & 0.63 & 22.76 \\
    \bottomrule
    \end{tabular}
\end{minipage}
\vspace{1em}

\subsubsection{Generation Ability}
\begin{table*}[t]
\caption{\textbf{Class-conditional generation performance on ImageNet 256×256.} DecQ achieves superior generation quality compared to various types of tokenizers, both with and without guidance. These results suggest that DecQ provides strong generative modeling capability.}
\centering
\label{tab:gen_main}
\resizebox{\textwidth}{!}{%
\begin{tabular}{l cc cccc cccc}
\toprule
\multirow{2}{*}{\textbf{Method}} & 
\multirow{2}{*}{\textbf{Epochs}} & 
\multirow{2}{*}{\textbf{\#Params}} & 
\multicolumn{4}{c}{\textbf{Generation w/o guidance}} & 
\multicolumn{4}{c}{\textbf{Generation w/ guidance}} \\
\cmidrule(lr){4-7} \cmidrule(lr){8-11}
& & & 
\textbf{gFID$\downarrow$} & \textbf{IS$\uparrow$} & \textbf{Pre.$\uparrow$} & \textbf{Rec.$\uparrow$} & 
\textbf{gFID$\downarrow$} & \textbf{IS$\uparrow$} & \textbf{Pre.$\uparrow$} & \textbf{Rec.$\uparrow$} \\
\midrule

\addlinespace[2pt]
\multicolumn{11}{c}{\textit{Traditional VAE Tokenizer}} \\
\addlinespace[2pt]

SiT \cite{SiT}        & 1400 & 675M    & 8.61 & 131.7 & 0.68 & \textbf{0.67} & 2.06 & 270.3 & \textbf{0.82} & 0.59 \\
REPA \cite{repa}       & 800  & 675M    & 5.90 & --    & --   & --   & 1.42 & \textbf{305.7} & 0.80 & 0.65 \\
MaskDiT \cite{maskdit}    & 1600 & 675M    & 5.69 & 177.9 & 0.74 & 0.60 & 2.28 & 276.6 & 0.80 & 0.61 \\
REG \cite{reg}       & 480  & 677M    & 2.2 & --    & --   & --   & 1.40 & 296.9 & 0.77 & 0.66 \\
\midrule
\addlinespace[4pt]
\multicolumn{11}{c}{\textit{VFM-Aligned VAE Tokenizer}} \\
\addlinespace[2pt]

VAVAE \cite{va-vae}      & 64  & 675M  & 5.14 & 130.2 & 0.76 & 0.62 & 2.11 & 252.3 & 0.81 & 0.58 \\
            & 800 & 675M  & 2.17 & 205.6 & 0.77 & 0.65 & 1.35 & 295.3 & 0.79 & 0.65 \\
\addlinespace[2pt]
DMVAE \cite{dmvae}      & 64  & 675M  & 3.22 & 171.7 & --   & --   & --   & --    & --   & --   \\
            & 800 & 675M  & 1.64 & 216.3 & --   & --   & --   & --    & --   & --   \\
\midrule
\addlinespace[4pt]
\multicolumn{11}{c}{\textit{Low-dim VFM Tokenizer}} \\
\addlinespace[2pt]

FAE \cite{fae}        & 80  & 675M  & 2.08 & 207.6 & \textbf{0.82} & 0.59 & 1.70 & 243.8 & \textbf{0.82} & 0.61 \\
            & 800 & 675M  & 1.48 & 239.8 & 0.81 & 0.63 & 1.29 & 268.0 & 0.80 & 0.64 \\
\addlinespace[2pt]
RPiAE \cite{rpiae}      & 60  & 675M  & 2.46 & 201.1 & 0.80 & 0.59 & 2.06 & 208.5 & 0.80 & 0.61 \\
            & 80  & 675M  & 2.25 & 208.7 & 0.81 & 0.60 & 1.51 & 225.9 & 0.79 & 0.65 \\
\midrule
\addlinespace[4pt]
\multicolumn{11}{c}{\textit{High-dim VFM Tokenizer}} \\
\addlinespace[2pt]

RAE \cite{rae}        & 80  & 839M  & 2.16 & 214.8 & \textbf{0.82} & 0.59 & --   & --    & --   & --   \\
            & 800 & 839M  & 1.51 & 242.9 & 0.79 & 0.63 & 1.13 & 262.6 & 0.78 & \textbf{0.67} \\
\addlinespace[2pt]
LV-RAE \cite{lvrae}     & 800 & 839M  & 2.42 & 223.8 & 0.77 & 0.64 & 1.82 & 249.7 & 0.75 & \textbf{0.67} \\
\addlinespace[2pt]
DecQ  & 80  & 841M  & 1.80 & 223.9 & 0.81 & 0.61 & 1.33   & 234.3    & 0.80   & 0.64   \\
            & 800 & 841M  & \textbf{1.41} & \textbf{251.9} & 0.81 & 0.63 & \textbf{1.05} & 259.6 & 0.79 & 0.66 \\

\bottomrule
\end{tabular}%
}
\end{table*}

We report the main generation results in \cref{tab:gen_main}. Existing LDM-based methods can be broadly categorized into four groups: (1) traditional approaches based on standard VAEs, (2) methods that enhance VAEs with semantic alignment, (3) methods that employ VFMs as tokenizers and perform generation in a low-dimensional latent space, and (4) methods that directly generate in the high-dimensional VFM feature space. 
For high-dimensional generation, DecQ follows RAE and adopts the same generative architecture and training settings. Additional implementation details are provided in Appendix~\ref{sec:implement}. 
Experimental results show that DecQ achieves an FID of 1.80 at 80 epochs and 1.41 at 800 epochs without guidance, and further improves to 1.05 at 800 epochs with guidance, outperforming previous state-of-the-art methods. Additional sampling details and qualitative generation results are provided in Appendix~\ref{sec:qualitative}.

\subsection{Ablation Study}
\label{sec:ablation}

\begin{table*}[t]
\caption{\textbf{Performance comparison of different VFM-based tokenizer training frameworks.}}%
\centering
\small
\label{tab:backbone_ablation}
\begin{tabular}{l ccc cccc} %
\toprule
\multirow{2}{*}{\textbf{Method}} & 
\multirow{2}{*}{\textbf{rFID$\downarrow$}} & 
\multirow{2}{*}{\textbf{SSIM$\uparrow$}} & 
\multirow{2}{*}{\textbf{PSNR$\uparrow$}} & 
\multicolumn{4}{c}{\textbf{w/o guidance 80 epoch}} \\ %
\cmidrule(lr){5-8} %
& & & & \textbf{FID$\downarrow$} & \textbf{IS$\uparrow$} & \textbf{Pre.$\uparrow$} & \textbf{Rec.$\uparrow$} \\ %
\midrule
VFM-freeze (RAE)   & 0.67 & 0.51 & 19.61 & 6.05 & 146.6 & \textbf{0.80} & 0.53 \\ 
VFM-finetune   & \textbf{0.15} & \textbf{0.94} & \textbf{33.83} & 57.81 & 26.32 & 0.40 & 0.56 \\ 
VFM-distill    & 0.23 & 0.81 & 27.09 & 20.20 & 81.02 & 0.62 & 0.57 \\ 
VFM-feat-concat  & 0.49 & 0.61 & 22.60 & 8.75  & 131.1 & 0.75 & 0.52 \\ 
DecQ          & 0.47 & 0.63 & 22.76 & \textbf{4.74}  & \textbf{148.9} & 0.78 & \textbf{0.59} \\ 

\bottomrule
\end{tabular}
\end{table*}

In \cref{tab:backbone_ablation}, we compare different backbone training paradigms, including freezing the VFM (RAE), unfreezing with and without distillation, feature concatenation, and our proposed DecQ. For distillation, we use an $L_2$ loss to align the encoder outputs with those of a frozen copy of the VFM. For the feature concatenation baseline, we set the number of query tokens equal to the number of patch tokens, train a low-dimensional bottleneck during reconstruction, and concatenate the resulting query features with patch tokens to form a new latent space.

Overall, the results reveal a clear reconstruction--generation trade-off. Freezing the VFM preserves strong generative performance but limits reconstruction, while full fine-tuning substantially improves reconstruction at the cost of degraded generation. Adding distillation slightly alleviates this issue but does not resolve the trade-off. Feature concatenation improves reconstruction but still underperforms in generation, suggesting that directly concatenating query features with patch tokens does not necessarily yield a well-aligned latent space for generative modeling. In contrast, DecQ improves both reconstruction and generation by preserving the original semantic structure while augmenting it with complementary fine-grained information, effectively mitigating the trade-off.

\begin{figure}[h]%
\centering
\vspace{-2mm}
\begin{minipage}[h]{0.49\textwidth}
    In \cref{fig:conv}, we provide a comparison evaluating the convergence behavior and training efficiency of our proposed DecQ framework with REPA \cite{repa} and RAE \cite{rae}. The evaluation is conducted using the FID-50K metric on the ImageNet dataset at the resolution of 256x256. DecQ exhibits an accelerated convergence trajectory compared to both REPA and RAE from the earliest stages of training. Specifically, DecQ achieves a gFID of 1.80 after only 80 epochs, and further improves to 1.51 at 240 epochs, matching the performance of RAE trained for 800 epochs. This corresponds to a 3.3$\times$ faster convergence rate, demonstrating that DecQ enables faster generative modeling.
\end{minipage}
\hfill
\begin{minipage}[h]{0.48\textwidth}
    \centering
    \includegraphics[height=4.4cm]{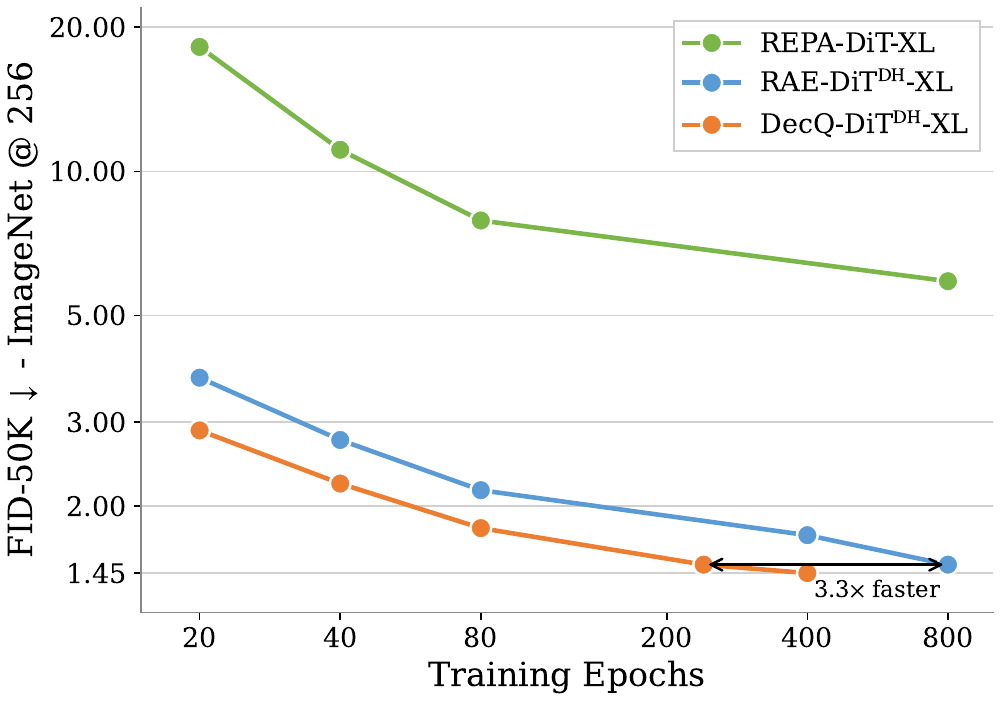}
    \caption{
        \textbf{Convergence of the proposed DecQ compared with REPA \cite{repa} and RAE \cite{rae}.}
    }
    \label{fig:conv}
\end{minipage}
\vspace{-3mm}
\end{figure}

\begin{wraptable}{r}{0.48\textwidth} %
\centering
\vspace{-3mm}
\caption{\textbf{Performance comparison of RAE, DecQ and DecQ (RAE decoder).}}
\label{tab:woquery_ablation}
\small
\begin{tabular}{l cc}
\toprule
\multirow{2}{*}{\textbf{Method}} & \multicolumn{2}{c}{\textbf{w/o guidance 80 epoch}} \\
\cmidrule(lr){2-3}
& \textbf{FID$\downarrow$} & \textbf{IS$\uparrow$} \\
\midrule
RAE              & 2.16 & 214.8 \\
DecQ (RAE decoder) & 1.99 & \textbf{229.8} \\
DecQ            & \textbf{1.80} & 223.9 \\
\bottomrule
\end{tabular}
\vspace{-2mm}
\end{wraptable}

In \cref{tab:woquery_ablation}, we compare RAE, DecQ, and DecQ (RAE decoder). 
DecQ (RAE decoder) uses the DecQ tokenizer for diffusion training, but discards the generated query tokens at inference and decodes only the generated patch tokens with the RAE decoder. 
Since DecQ preserves the original VFM patch-token latent space, replacing the DecQ decoder with the RAE decoder does not introduce a latent-space mismatch for the patch tokens. 
Interestingly, DecQ (RAE decoder) still outperforms RAE even when the generated query tokens are discarded at inference. This suggests that predicting detail-condensing queries may itself help the diffusion model generate better patch tokens, in a way reminiscent of REG~\cite{reg}.
Moreover, the full DecQ model further improves over DecQ (RAE decoder), showing that the generated query tokens carry fine-grained information that directly benefits decoding and final generation quality.

\begin{table}[t]
\caption{\textbf{Ablation on the number of queries.} More queries improve reconstruction but do not always benefit generation. Using 8 queries provides the best reconstruction-generation trade-off.}
\centering
\label{tab:querynum_ablation}
\resizebox{\textwidth}{!}{%
\small
\begin{tabular}{l ccc cc cc}
\toprule
\multirow{2}{*}{\textbf{Number of queries}} & 
\multirow{2}{*}{\textbf{rFID$\downarrow$}} & 
\multirow{2}{*}{\textbf{SSIM$\uparrow$}} & 
\multirow{2}{*}{\textbf{PSNR$\uparrow$}} & 
\multicolumn{2}{c}{\textbf{w/o guidance 80 epoch}} & 
\multicolumn{2}{c}{\textbf{Additional overhead}} \\
\cmidrule(lr){5-6} \cmidrule(lr){7-8}
& & & & \textbf{FID$\downarrow$} & \textbf{IS$\uparrow$} & \textbf{Flops} & \textbf{Params} \\
\midrule

RAE & 0.67 & 0.51 & 19.61 & 6.05 & 146.6 & -- & -- \\
\midrule

$K=2$   & 0.57 & 0.54 & 20.89 & 5.94 & 146.9 & +1.7\% & +29.3\,M \\
$K=4$   & 0.54 & 0.56 & 21.33 & 5.22 & \textbf{151.1} & +2.4\% & +29.3\,M \\
$K=8$   & 0.47 & 0.63 & 22.76 & \textbf{4.74} & 148.9 & +3.9\% & +29.3\,M \\
$K=16$  & \textbf{0.32} & \textbf{0.71} & \textbf{24.41} & 6.43 & 138.7 & +6.8\% & +29.3\,M \\

\bottomrule
\end{tabular}%
}
\end{table}

We study the effect of varying the number of queries in \cref{tab:querynum_ablation}. Increasing the number of queries consistently improves reconstruction, as additional queries can condense richer low-level information from intermediate VFM features. However, better reconstruction does not always translate to better generation. With DiT$^{\mathrm{DH}}$-S, FID first decreases and then increases as more queries are used, suggesting that a moderate number of queries provides useful complementary details, whereas excessive queries may introduce redundant low-level information that interferes with generative modeling. In practice, using 8 queries achieves the best generative performance, striking a favorable balance between reconstruction fidelity and generation quality. Overhead analysis is provided in Appendix~\ref{sec:overhead}.

\begin{table}[t]
\caption{\textbf{Ablation on condenser placement across VFM layers.} Shallow layers favor reconstruction, while deeper layers benefit generation. The sparse configuration at layers 0, 3, 6, and 9 achieves a balanced trade-off between fidelity, quality, and efficiency.}
\centering
\label{tab:querylayer_ablation}
\resizebox{\textwidth}{!}{%
\small
\begin{tabular}{l ccc cc cc}
\toprule
\multirow{2}{*}{\textbf{Layers}} & 
\multirow{2}{*}{\textbf{rFID$\downarrow$}} & 
\multirow{2}{*}{\textbf{SSIM$\uparrow$}} & 
\multirow{2}{*}{\textbf{PSNR$\uparrow$}} & 
\multicolumn{2}{c}{\textbf{w/o guidance 80 epoch}} & 
\multicolumn{2}{c}{\textbf{Additional overhead}} \\
\cmidrule(lr){5-6} \cmidrule(lr){7-8}
& & & & \textbf{FID$\downarrow$} & \textbf{IS$\uparrow$} & \textbf{Flops} & \textbf{Params} \\
\midrule

RAE       & 0.67 & 0.51 & 19.61 & 6.05 & 146.6 & -- & -- \\
\midrule

Layer 0,1,2,3   & \textbf{0.42} & \textbf{0.66} & \textbf{23.47} & 6.35 & 143.5 & +3.9\% & +29.3\,M \\
Layer 8,9,10,11 & 0.54 & 0.54 & 20.72 & 5.09 & \textbf{156.1} & +3.9\% & +29.3\,M \\
All layers      & 0.45 & 0.62 & 22.67 & 4.77 & 150.5 & +6.1\% & +86.1\,M \\
Layer 0,3,6,9   & 0.47 & 0.63 & 22.76 & \textbf{4.74} & 148.9 & +3.9\% & +29.3\,M \\

\bottomrule
\end{tabular}%
}
\end{table}

We study the effect of applying condensers at different VFM layers in \cref{tab:querylayer_ablation}. Different depths show distinct behaviors: shallow layers provide richer low-level details and improve reconstruction, but degrade generative performance; deeper layers offer better generation by capturing higher-level semantics. This reveals a layer-dependent reconstruction--generation trade-off.
While applying condensers to all layers achieves strong performance, it also introduces higher computational and parameter overhead. We therefore adopt a sparse design at layers 0, 3, 6, and 9, which performs comparably to dense aggregation while better balancing reconstruction fidelity, generation quality, and computational cost.

\begin{figure}[h] %
\centering
\vspace{-3mm}
\begin{minipage}[h]{0.59\textwidth}
    We analyze the roles of query and patch tokens through a clustering-based study. Using the top-left image as the anchor, we retrieve its nearest neighbors under query-token and patch-token representations, as shown in \cref{fig:cluster}. Query-token clusters tend to share color-related visual patterns with the anchor, suggesting that query tokens primarily capture low-level appearance details such as color and texture. In contrast, patch-token clusters consistently retrieve images with similar semantic content and object-level structures, indicating stronger high-level category information. This qualitative comparison highlights the complementary roles of the two representations: query tokens enrich fine-grained visual details, while patch tokens preserve semantic structure. More results are provided in Appendix~\ref{sec:qualitative}.%
    
\end{minipage}
\hfill
\begin{minipage}[h]{0.4\textwidth}
    \centering
    \includegraphics[height=4.5cm]{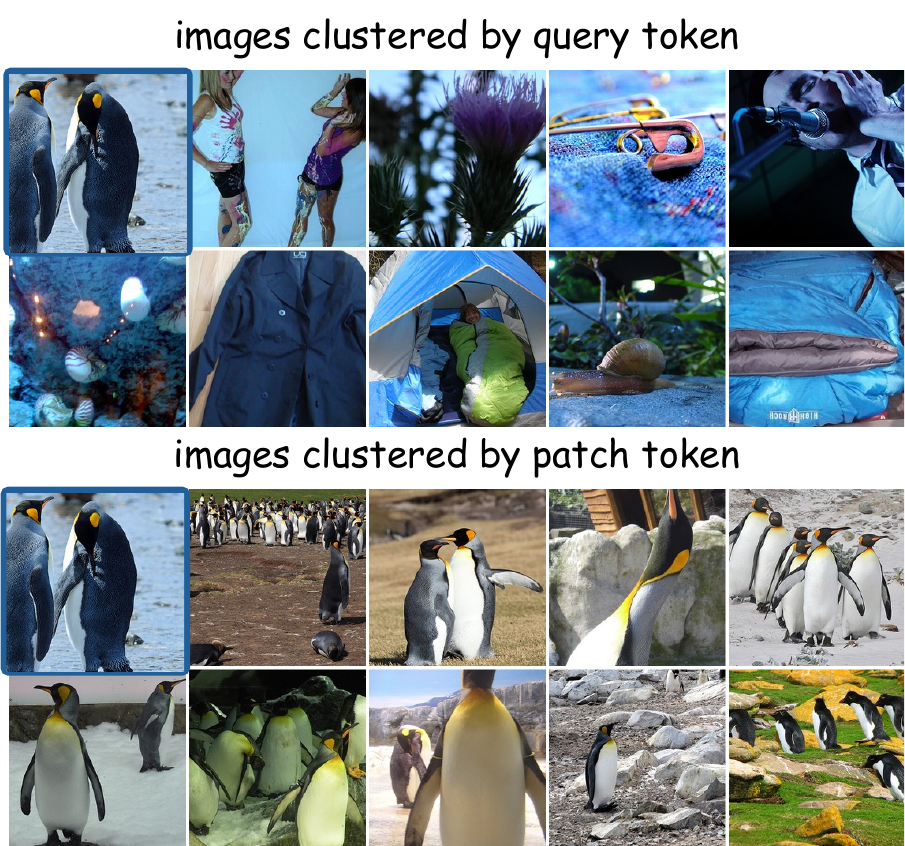}
    \caption{
        \textbf{Images clustered using different token representations.} %
    }
    \label{fig:cluster}
\end{minipage}
\vspace{-3mm}
\end{figure}

\begin{table}[t]
\caption{\textbf{Performance comparison of RAE and DecQ based on SigLIP2.} Results show that DecQ remains effective across different VFMs, highlighting its robustness and general applicability.}
\centering
\small
\label{tab:siglip2_ablation}
\resizebox{\textwidth}{!}{%
\begin{tabular}{l ccc cc cc}
\toprule
\multirow{2}{*}{\textbf{Method}} & 
\multirow{2}{*}{\textbf{rFID$\downarrow$}} & 
\multirow{2}{*}{\textbf{SSIM$\uparrow$}} & 
\multirow{2}{*}{\textbf{PSNR$\uparrow$}} & 
\multicolumn{2}{c}{\textbf{w/o guidance 80 epoch}} & 
\multicolumn{2}{c}{\textbf{Additional overhead}} \\
\cmidrule(lr){5-6} \cmidrule(lr){7-8}
& & & & \textbf{FID$\downarrow$} & \textbf{IS$\uparrow$} & \textbf{Flops} & \textbf{Params} \\
\midrule

RAE   & 0.73 & 0.51 & 19.92 & 11.10 & 107.7 & -- & -- \\
DecQ  & \textbf{0.57} & \textbf{0.60} & \textbf{22.07} & \textbf{10.09} & \textbf{111.5} & +3.9\% & +29.3M \\

\bottomrule
\end{tabular}%
}
\end{table}

\paragraph{Generalization across different VFMs.} 
To evaluate generality, we conduct analogous experiments with SigLIP2-B, as reported in \cref{tab:siglip2_ablation}. Consistent with DINOv2, reconstruction with a frozen SigLIP2 is limited, while introducing detail-condensing queries substantially improves pixel-level metrics. Although SigLIP2 shows lower reconstruction and generative performance than DINOv2, DecQ still brings consistent gains under the same setting. These results show that DecQ remains effective across different VFM architectures, highlighting its robustness and general applicability.

\section{Conclusion}

We presented DecQ, a framework that introduces detail-condensing queries to attend to intermediate VFM layers via cross-attention, recovering fine-grained information progressively lost in VFM representations. DecQ enriches low-level details while preserving the VFM semantic latent space. During generation, it jointly denoises patch and query tokens, enabling richer details and higher generation quality. Experiments consistently show that DecQ improves both reconstruction and generation with minimal computational overhead.

\bibliographystyle{unsrt}
\bibliography{main}

\newpage
\appendix

\section{Implementation Details}
\label{sec:implement}

\subsection{DecQ Implementation}

We follow the training scheme of RAE. For the encoder, we use DINOv2 with Registers~\cite{registers} to process images resized to $224 \times 224$, producing 256 patch tokens that are then used to reconstruct images at $256 \times 256$ resolution. Both the [CLS] and [REG] tokens are discarded after encoding. For both patch tokens and the newly introduced query tokens, we apply layer normalization with \texttt{elementwise\_affine=False} to ensure proper normalization. We adopt the same noise injection strategy as RAE for patch tokens, and apply noise with the same variance to the query tokens.

We follow the decoder design of RAE. The query tokens are projected independently, assigned learnable positional embeddings, and concatenated with patch tokens to form a unified input sequence. The combined sequence is processed jointly by the Transformer decoder. Only patch tokens are used for final reconstruction, while query tokens serve as auxiliary latent variables that enhance the decoding process. For experiments with SigLIP2, we use the same DecQ configuration as DINOv2 unless otherwise specified.

\subsection{Diffusion Model Implementation}

Following RAE, we use LightningDiT~\cite{va-vae} as the backbone of our diffusion model. We adopt a continuous-time flow matching formulation, where the timestep is defined over the real interval $[0, 1]$, and replace the standard timestep embedding with Gaussian Fourier feature embeddings.

For DiT$^{\mathrm{DH}}$, we largely follow RAE, using DiT$^{\mathrm{DH}}$-XL for the main results and DiT$^{\mathrm{DH}}$-S for ablations. When the hidden dimension of the DiT backbone differs from that of the DDT head, we use a linear projection layer to map the encoder output to the decoder dimension.

For optimization, we mainly follow the LightningDiT training recipe. We use AdamW with a constant learning rate of $2.0 \times 10^{-4}$, a batch size of 1024, and an EMA decay of 0.9999. We also apply gradient clipping with a threshold of 1.0. The query-token loss weight $\lambda_{\mathrm{query}}$ is set to 1, since query and patch tokens are constrained to have the same variance during tokenizer training. All diffusion models are trained on 8 NVIDIA H200 GPUs.

\subsection{Sampling Details}

We use standard ODE sampling with an Euler solver and default to 50 sampling steps. We also observe that increasing the number of steps to 250 can yield further improvements. For FID-50K evaluation, we follow the RAE protocol and sample 50 images per class, resulting in 50,000 images in total.

Following RAE, we adopt AutoGuidance~\cite{AG} as our primary guidance strategy, which uses a weaker diffusion model to guide a stronger one. Consistent with RAE, we use the minimal variant, DiT$^{\mathrm{DH}}$-S, as the guiding model, initialized from a relatively early checkpoint. Our best results are obtained using the 60-epoch checkpoint of DiT$^{\mathrm{DH}}$-S with a guidance scale of 1.6.

\section{More Qualitative Results}
\label{sec:qualitative}
We present additional qualitative results, including clustering analyses that illustrate the role of query tokens, reconstruction comparisons on DINOv2 and SigLIP2, and qualitative generated samples.

\subsection{Cluster Analysis}
\begin{figure}[t]
  \centering

  \begin{minipage}{0.48\textwidth}
    \centering
    \includegraphics[width=\linewidth]{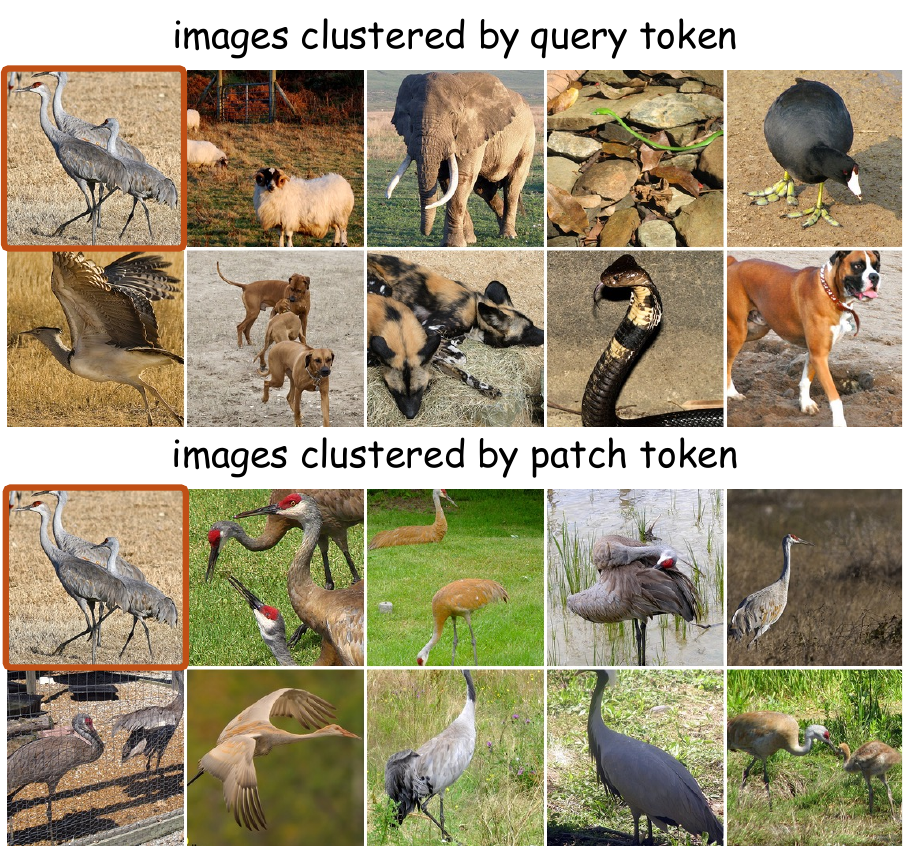}
    \label{fig:cluster1}
  \end{minipage}
  \hfill
  \begin{minipage}{0.48\textwidth}
    \centering
    \includegraphics[width=\linewidth]{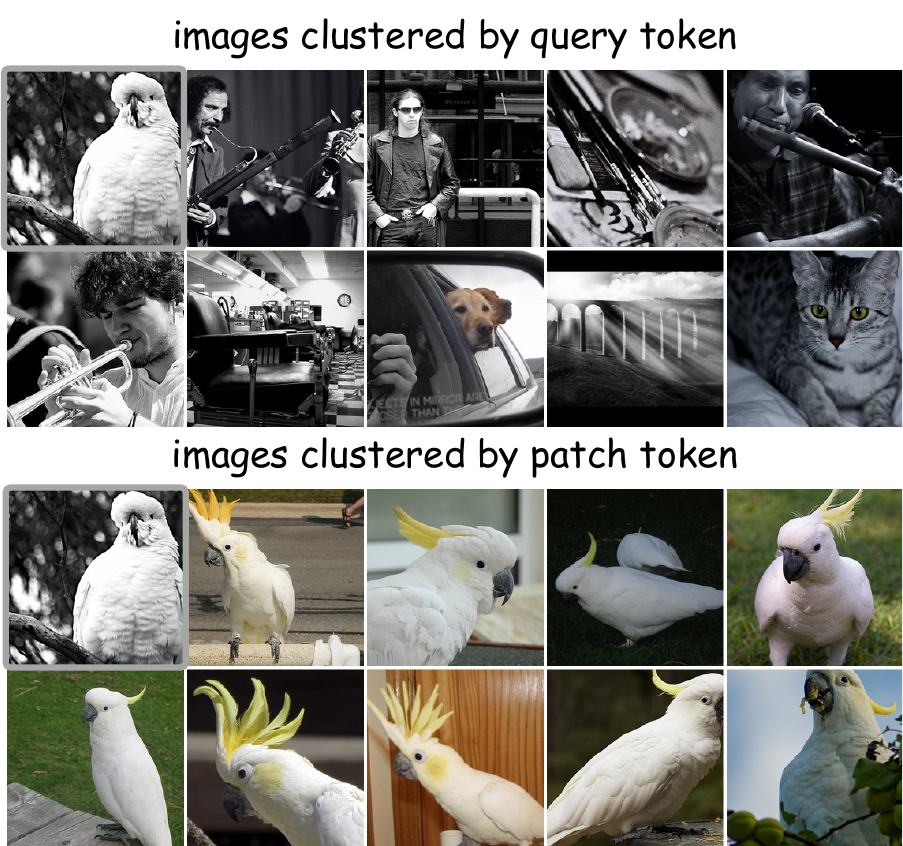}
    \label{fig:cluster2}
  \end{minipage}

  \caption{
    \textbf{More results on cluster visualization.} Similar to \cref{fig:cluster}, images clustered by query tokens shares similar appearances like background color, while images clustered by patch tokens share high-level semantics such as main subjects. This suggests that query tokens mainly capture fine-grained visual details, while patch tokens preserve high-level semantics.
  }
  \label{fig:clusters}
\end{figure}

Additional clustering results are shown in \cref{fig:clusters}. They further show that query tokens primarily capture fine-grained visual details such as color and texture, while patch tokens preserve high-level semantics such as object identity.

\subsection{Reconstruction Performance}

\begin{figure}[t]
  \centering
  \includegraphics[width=\textwidth]{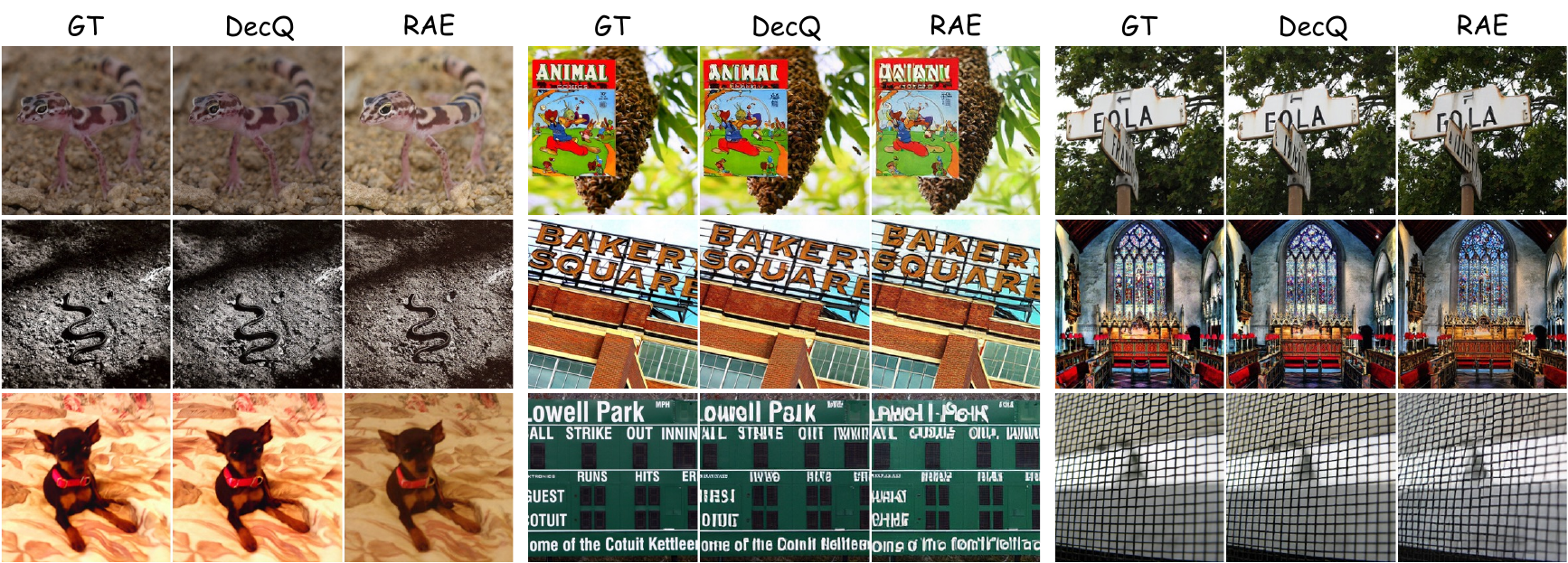}
  \caption{
        \textbf{More qualitative results of our image reconstruction compared with RAE based on DINOv2.} The cases share the same setting as in \cref{fig:teaser} Right. Compared with RAE, DecQ better preserves background colors, textual content, and fine-grained textures.
  }
  \label{fig:recon_appendix}
\end{figure}

Additional reconstruction comparisons with the DINOv2-based RAE are shown in \cref{fig:recon_appendix}, using the same setting as in \cref{fig:teaser} Right.

\begin{figure}[t]
  \centering
  \includegraphics[width=\textwidth]{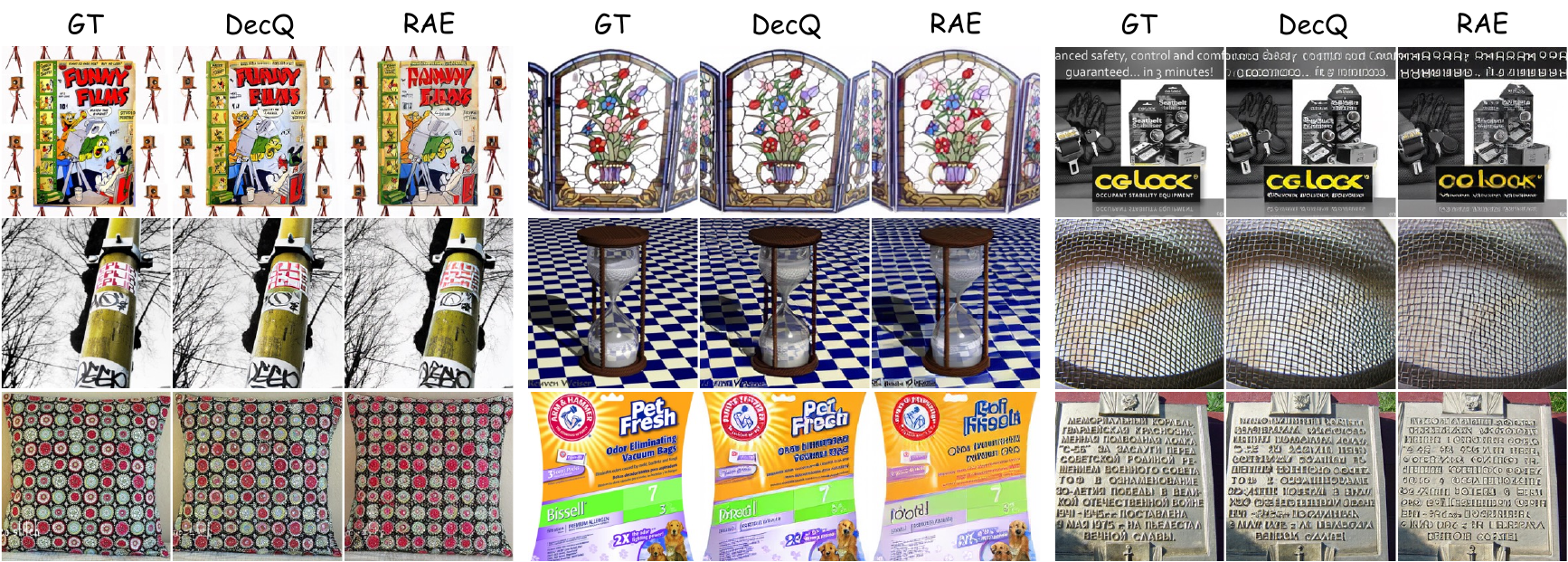}
  \caption{
        \textbf{Qualitative results of our image reconstruction compared with RAE based on SigLIP2.} In some cases, SigLIP2 appears to retain a semantic impression of textual content but fails to accurately reproduce its colors. In contrast, DecQ better preserves these fine-grained details.
  }
  \label{fig:recon_siglip}
\end{figure}

Reconstruction results compared with RAE based on SigLIP2 are presented in \cref{fig:recon_siglip}.
In some cases, the SigLIP2-based RAE retains a semantic impression of textual content but fails to faithfully reproduce its colors, whereas DecQ better preserves these fine-grained details.

\subsection{Generation Performance}
\begin{figure}[t]
  \centering
  \includegraphics[width=\textwidth]{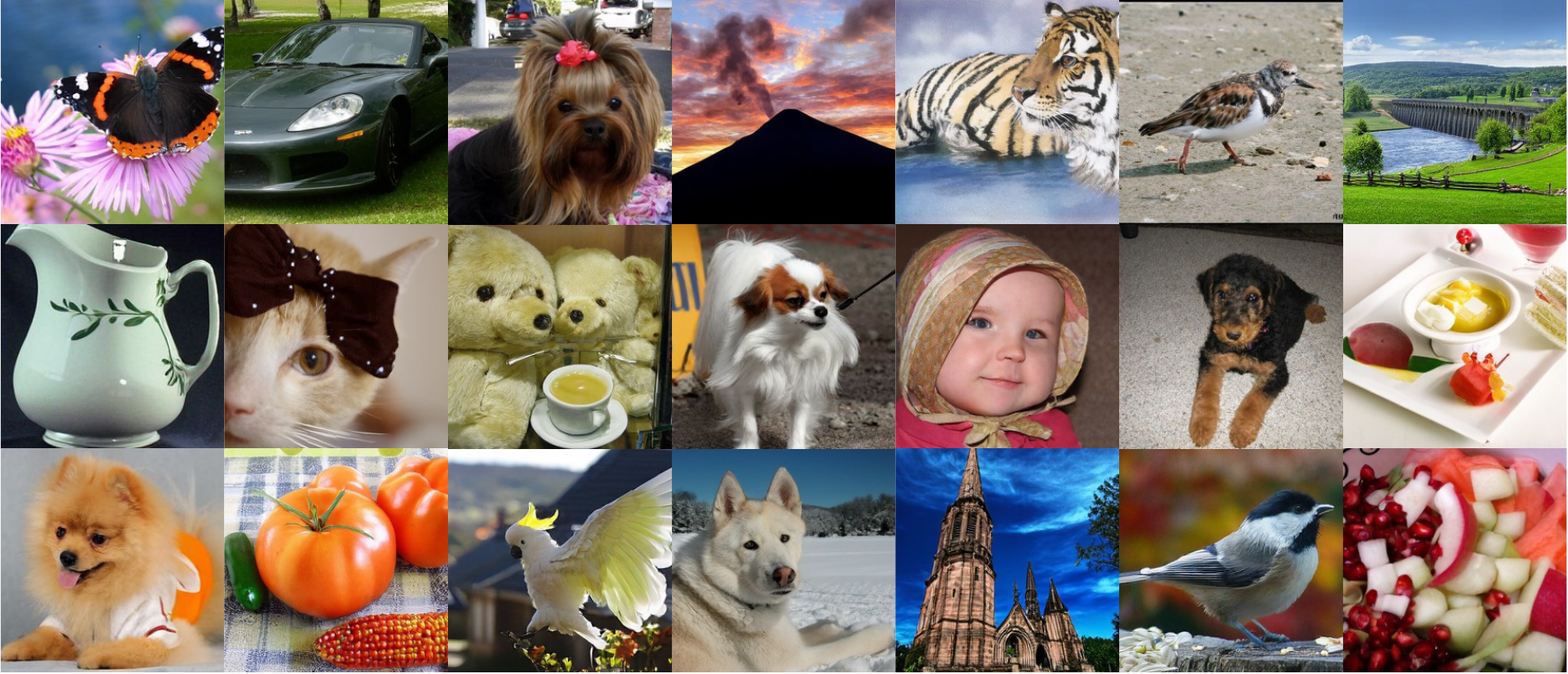}
  \caption{
        \textbf{Qualitative results of our image generation.}
  }
  \label{fig:gen_result}
\end{figure}

Our class-to-image generation results are presented in \cref{fig:gen_result}, demonstrating the strong generative capabilities of DecQ. %

\section{Computational Overhead Analysis}
\label{sec:overhead}

We analyze the additional computational and parameter overhead introduced by DecQ. 
All GFLOPs are reported under the MACs convention, following RAE~\cite{rae}.

\subsection{Tokenizer and Reconstruction Overhead}

\paragraph{Baseline.}
The frozen ViT-B/14 encoder processes $N{=}256$ tokens through 12 Transformer layers ($d{=}768$, $d_{\mathrm{ff}}{=}3072$), resulting in 22.2~GFLOPs. 
The ViT-MAE XL decoder processes $N_{\mathrm{dec}}{=}257$ tokens through 28 layers ($d_{\mathrm{dec}}{=}1152$, $d_{\mathrm{ff}}{=}4096$), costing 106.7~GFLOPs. 
This gives a total baseline cost of 128.9~GFLOPs per image, with 501.9M active parameters.

\paragraph{Computational overhead.}
DecQ introduces two sources of additional computation. 
First, DecQ inserts $M{=}4$ condensers, where $M$ denotes the number of VFM layers equipped with a condenser, to extract $K{=}8$ query tokens, adding 1.44~GFLOPs.
Second, concatenating the query tokens increases the decoder sequence length from 257 to 265, adding 3.58~GFLOPs. 
In total, the default query mechanism adds 5.0~GFLOPs, corresponding to only +3.9\% over the baseline.

\paragraph{Parameter overhead.}
Each condenser contains a cross-attention extractor and a feed-forward network, contributing 7.09M parameters per condenser.
With $M{=}4$ condensers, this amounts to 28.36M parameters.
The remaining components, including learnable queries, the decoder query projection, and query positional embeddings, contribute only a minor overhead. 
Overall, DecQ introduces 29.3M additional trainable parameters, corresponding to +5.8\% of the 501.9M-parameter baseline. 
The parameter overhead is governed mainly by the condenser dimension and the number of inserted condensers, and is nearly independent of the number of query tokens $K$.

\paragraph{Summary.}
Table~\ref{tab:overhead} summarizes the overhead under different configurations. 
The default setting ($M{=}4$, $K{=}8$) adds only +3.9\% computation and +5.8\% parameters, showing that DecQ improves the tokenizer with modest additional cost.

\begin{table}[t]
\centering
\caption{Tokenizer and reconstruction overhead of DecQ. $M$ denotes the number of condenser-equipped VFM layers, and $K$ denotes the number of query tokens. GFLOPs follow the MACs convention. ``Active Params'' counts parameters that participate in forward computation.}
\label{tab:overhead}
\begin{tabular}{lccccc}
\toprule
Configuration & \multicolumn{3}{c}{GFLOPs} & Active Params & Overhead \\
\cmidrule(lr){2-4}
($M \times K$) & Encoder & Decoder & Total & ($\Delta$) & (FLOPs / Params) \\
\midrule
Baseline (no query) & 22.2 & 106.7 & 128.9 & --- & --- \\
\midrule
$4 \times 2$  & 23.5 & 107.6 & 131.1 & +29.3\,M & +1.7\% / +5.8\% \\
$4 \times 4$  & 23.5 & 108.5 & 132.0 & +29.3\,M & +2.4\% / +5.8\% \\
$4 \times 8$  & 23.6 & 110.3 & 133.9 & +29.3\,M & +3.9\% / +5.8\% \\
$4 \times 16$ & 23.8 & 113.9 & 137.7 & +29.3\,M & +6.8\% / +5.8\% \\
$4 \times 32$ & 24.2 & 121.1 & 145.3 & +29.3\,M & +12.7\% / +5.8\% \\
$12 \times 8$ & 26.5 & 110.3 & 136.8 & +86.1\,M & +6.1\% / +17.1\% \\
\bottomrule
\end{tabular}
\end{table}

\subsection{Generation Overhead}

We further analyze the overhead introduced by query tokens during diffusion generation with DiT$^{\mathrm{DH}}$-XL. 
The baseline generation pipeline consists of 50 ODE sampling steps followed by one RAE decoding step, resulting in 8,189.8~GFLOPs in total. 
For the default setting $K{=}8$, appending query tokens increases the DiT cost by 5.20~GFLOPs per sampling step. 
Together with the additional 3.58~GFLOPs in the final RAE decoder, the complete generation overhead is 263.4~GFLOPs, corresponding to +3.22\% over the baseline.

The parameter overhead during generation is even smaller. 
Unlike the tokenizer stage, which requires cross-attention condensers, the generation stage only introduces lightweight embedding and projection layers for the additional query tokens.
For $K{=}8$, this adds 3.37M parameters in total. 
Since most of these parameters come from embedding projections rather than query-specific positional embeddings, the parameter overhead is nearly independent of $K$.

Table~\ref{tab:generation_overhead} summarizes the generation overhead for different numbers of query tokens. 
The additional computation scales approximately linearly with $K$, while the quadratic attention term remains negligible in this regime. 
Overall, DecQ introduces only a modest overhead during generation, with the default setting adding +3.22\% computation and 3.37M parameters.

\begin{table}[t]
\centering
\caption{Generation overhead of DecQ with DiT$^{\mathrm{DH}}$-XL. GFLOPs are computed for complete generation, including 50 ODE sampling steps and one RAE decoding step. The baseline cost is 8,189.8~GFLOPs.}
\label{tab:generation_overhead}
\begin{tabular}{cccc}
\toprule
Query tokens $K$ & DiT $\Delta$/step & Total GFLOPs $\Delta$ & Param. $\Delta$ \\
\midrule
2  & +1.30 & +65.8  (+0.80\%)  & +3.35\,M \\
4  & +2.60 & +131.6 (+1.61\%)  & +3.36\,M \\
8  & +5.20 & +263.4 (+3.22\%)  & +3.37\,M \\
16 & +10.40 & +527.2 (+6.44\%) & +3.39\,M \\
32 & +20.84 & +1056.3 (+12.90\%) & +3.42\,M \\
\bottomrule
\end{tabular}
\end{table}

\section{Limitations}
\label{sec:limitations}

While DecQ consistently improves both reconstruction and generation under our experimental settings, several limitations remain. First, our experiments are mainly conducted on ImageNet at $256 \times 256$ resolution. Although ImageNet is a standard benchmark for class-conditional image generation, evaluating DecQ on more diverse datasets, such as text-to-image datasets or domain-specific image collections, would provide a more comprehensive understanding of its generality. Second, we have not extensively studied higher-resolution generation, such as $512 \times 512$ or above. Since fine-grained details become increasingly important at higher resolutions, it would be valuable to examine how the number of queries, condenser placement, and computational overhead scale in such settings.

In addition, our method is evaluated primarily with DINOv2 and SigLIP2 as representative VFMs. While the results suggest that DecQ generalizes across different VFM architectures, a broader study covering more backbone families and model scales remains an important direction. Finally, DecQ introduces additional query tokens and condenser modules. Although the default configuration incurs only modest overhead, the cost may increase when more queries or denser condenser placements are used. Future work may explore adaptive query allocation or more efficient condenser designs to further reduce the overhead while preserving the reconstruction and generation benefits.

\end{document}